# Robust Planning in Uncertain Environments

**Stephen G. Pimentel**
Science Applications International Corporation
1710 Goodridge Drive
McLean, VA 22102
email: pimentel@sytex.com
Fax: 703-351-2629

**Lawrence M. Brem**
Adroit Systems, Inc.
209 Madison Street
Alexandria, VA 22314
email: lbrem@alexandria.adroit.com
Fax: 703-836-7411

## Abstract

This paper describes a novel approach to planning which takes advantage of decision theory to greatly improve robustness in an uncertain environment. We present an algorithm which computes conditional plans of maximum expected utility. This algorithm relies on a representation of the search space as an AND/OR tree and employs a depth-limit to control computation costs. A numeric robustness factor, which parameterizes the utility function, allows the user to modulate the degree of risk-aversion employed by the planner. Via a look-ahead search, the planning algorithm seeks to find an optimal plan using expected utility as its optimization criterion. We present experimental results obtained by applying our algorithm to a non-deterministic extension of the blocks world domain. Our results demonstrate that the robustness factor governs the degree of risk embodied in the conditional plans computed by our algorithm.

## 1 INTRODUCTION

This paper describes a novel approach to planning which takes advantage of decision theory to greatly improve the robustness of planning in an uncertain environment. The proposed concept improves upon conventional, goal-oriented planning techniques by the use of utility functions to guide the planner through a state space. The use of utility functions in planning permits tradeoffs between objectives and partial satisfaction of objectives, unlike the more coarse-grained goal-oriented approach. The proposed decision-theoretic planning framework models the results of actions, including the possibility of execution failure, through the use of probability functions. Therefore, it is able to manage uncertainty by seeking to maximize the *expected* utility of a plan, rather than the utility of a plan assuming successful execution.

The decision-theoretic framework we employ allows the robustness of plans to be defined in terms which are independent of specific domains and planning algorithms, but which can be readily applied to both. Conventional planning seeks to maximize the utility of a plan assuming the success of its actions. In contrast, our approach accounts for the possibility of action failure using a numeric *robustness factor*. The robustness factor parameterizes the utility function and allows the user to select its form, thereby modulating the degree of risk-aversion employed by the planner. Via a look-ahead search, the planning framework seeks to find an optimal plan, using expected utility as its optimization criterion.

The remainder of the paper is organized as follows. In section 2, we describe our formalism for robust planning under uncertainty. This is an extension of the STRIPS formalism incorporating probabilities and utilities. In section 3, we show how functions for expected utility can be parameterized by a robustness factor in the context of our planning formalism. In section 4, we present a planning algorithm which computes conditional plans of maximum expected utility. This algorithm relies on a representation of the search space as an AND/OR tree and employs depth-limitation to control computation costs. In section 5, we describe experimental results obtained by applying our planning algorithm within an extended, non-deterministic blocks world domain. Our results demonstrate that, as intended, the robustness factor governs the degree of risk embodied in the conditional plans computed by our algorithm. Section 6 describes the relationship between our approach and other work in the literature, and section 7 gives references.

## 2 FORMALISM FOR ROBUST PLANNING

Many formalisms have been devised to encode the information needed by planners. A planning formalism supplies the means of encoding planning *operators*, used to describe the actions of which the system is capable, and *facts*, used to describe states of the world. Although many formalisms of greater sophistication have been developed, the formalism of STRIPS [Fikes, 1971] is the simplest and most fundamental, in that the planning operators of more sophisticated systems are usually extensions of STRIPS-style operators. Therefore, in order to focus our effort on the issues of uncertainty and robustness, we will start with STRIPS operators and extend them as needed to handle the latter factors.

A STRIPS planning operator contains *preconditions*, an *add-list*, and a *delete-list*. An operator's preconditions give the facts that must hold in a state before the operator can be applied. If the operator is applied to a state, the add-list and delete-list are used to produce a new state by deleting all facts in the delete-list and adding all facts in the add-list. Figure 1 gives an example of a STRIPS operator.

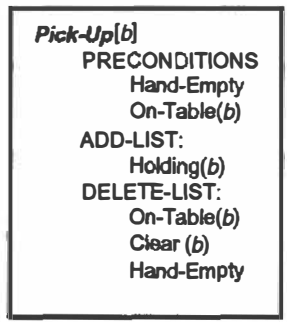

Figure 1. A STRIPS Planning Operator

The STRIPS formalism makes two key assumptions that will need to be relaxed in a decision-theoretic setting. First, it assumes that all operators succeed upon application. Consequently, the application of an operator can only result in exactly one new state, computed via the add-list and delete-list. In the real world, however, actions often fail to produce their desired result, and when they go wrong, they can do so in multiple ways. Hence, we will relax this assumption by allowing a planning operator to have $k$ different outcomes, for $k > 0$, each with its own add-list and delete-list. In other words, the application of an operator may result in any one of $k$ different new states. For each possible outcome, the operator will specify a probability function giving the likelihood of that outcome.

Second, STRIPS assumes that the planner is attempting to achieve a set of goals expressed as a conjunction of facts. This does not permit the planner to make tradeoffs between objectives or to seek partial satisfaction of objectives. Hence, we will extend planning operators with information expressing the increase or decrease of value for each their resulting states. Planning operators extended in this fashion will be called *decision operators*. Figure 2 gives an example of a decision operator. When a decision operator has exactly two possible outcomes, we may informally think of the higher-valued outcome as representing "success" for the action, and the lower-valued outcome as representing the "failure" of the action.

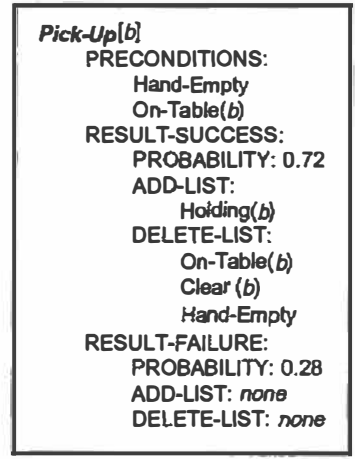

Figure 2. A Decision Operator

We can characterize the planning formalism more formally as follows. Every state $s$ in the domain is assumed to have a value, $v(s)$, which ranges between some $v_{min}$ and $v_{max}$, so that $v_{min} \leq v(s) \leq v_{max}$. However, it is more convenient to define utility functions on normalized values, so we also define a normalized value function

$$V(s) = \frac{v(s) - v_{min}}{v_{max} - v_{min}}$$

so that $0 \leq V(s) \leq 1$. It is important to note that the value function $V(s)$ is *not* a heuristic function of the sort used in A* and other search algorithms [Pearl, 1984]. Heuristic functions attempt to compute (a lower bound on) the "distance" between an intermediate state and some goal state. $V(s)$, on the other hand, computes the value of $s$ viewed as a (potential) terminal state. $V(s)$ is more closely related to the static evaluation functions used in minimax algorithms.

The application of a decision operator to a particular state is called an *action*. An action $a$ performed in a state $s$ is, in the terminology of decision theory, a lottery with $k$ mutually exclusive outcomes. Figure 3 depicts a lottery with $k = 3$.

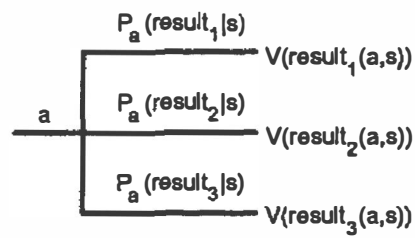

Figure 3. A Lottery with 3 Possible Outcomes.

We will denote the state corresponding to each outcome by $result_i(a,s)$ for $0 \leq i \leq k$. Each outcome occurs with probability $P_a(result_i|s)$ with the restriction that

$$\sum_{i=1}^{k} P_a(result_i|s) = 1$$

## 3 ROBUSTNESS-PARAMETERIZED EXPECTED UTILITY

A major advantage of our decision-theoretic framework is that it allows the robustness of plans to be defined in terms which are dependent on neither specific planning algorithms nor domains, but which can be readily translated into both. In our approach, robustness is a measure of a plan's capacity for "graceful degradation." It measures how well a plan performs when one or more actions within the plan fail. Our approach accounts for the possibility of action failure using a numeric robustness factor, R. In contrast, conventional, goal-oriented planners go to one end of the spectrum, seeking total efficacy. During planning, they discount the possibility of action failure, treating it strictly by replanning at execution-time.

Our planner employs a utility function to incorporate R into its evaluation of states and actions. The utility function will be characterized by a simple analytical form parameterized by R. In particular, it is a function $U_R(V)$ which is continuous on $0 \leq V \leq 1$ and undefined outside that range. The value of R is selected by the user as a system input. Figure 4 shows $U_R(V)$ for typical values of R, using the analytical form $U_R(V)=V^{1-R}$ with $0 \leq R < 1$.

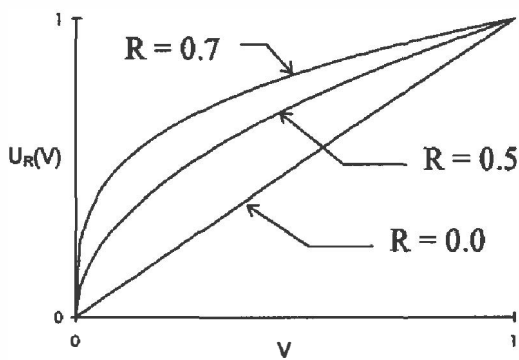

Figure 4. $U_R(V) = V^{(1-R)}$

A decision-theoretic planner strives to drive the world into a state of maximal utility through a sequence of actions. The planner will normally pass through a number of intermediate states before reaching a state of maximal utility. Therefore, the *expected* utility of an intermediate state is determined by the utility of the outcome to which we can expect it to ultimately lead, given a rational choice of further actions. A rational choice of actions, in turn, is one which maximizes the expected utility of the resulting states. These considerations leads to the following mutually-recursive pair of equations. A state will be called *terminal* if it is a possible final state for a planning scenario, and *non-terminal* otherwise. The expected utility of a state $s$ will be given by

$$EU_R(s) = \begin{cases} U_R(V(s)) & \text{if } s \text{ is terminal} \\ \max_a \{EU_R(a|s)\} & \text{if } s \text{ is non-terminal} \end{cases}$$

The expected utility of an action $a$ executed in a state $s$ will be given by

$$EU_R(a|s) = \sum_{i=1}^{k} P_a(result_i|s) EU_R(result_i(a,s))$$

if the preconditions of $a$ are satisfied in $s$, and will be 0 otherwise.

## 4 PLANNING ALGORITHM

The search space defined by our formalism can be described as an AND/OR tree [Nilsson, 1980] in which every state is an OR node and every action an AND node. From this perspective, the above equations can be readily translated into a recursive, look-ahead algorithm which, starting with an initial state $s$, searches the AND/OR tree down to the terminal states. The expected utilities will then be backed-up through the AND/OR tree, until returning to $s$. Each time the algorithm computes the expected utility of a node, it *solves* that node within the AND/OR tree. To solve an action (AND node), the algorithm computes the expected utility of all of its results and combines them according the equation for $EU_R(a|s)$. To solve a state (OR node), the algorithm selects the action which achieved the greatest expected utility of all the possible actions. An AND/OR tree is said to be solved when its root is solved. Since actions can have multiple possible outcomes, a solved AND/OR tree will have built-in contingencies for each of them. Hence, we will refer to a solved AND/OR tree as a *conditional plan*. Each path through a conditional plan corresponds to a traditional "linear" STRIPS plan. The planning algorithm integrates the expected utility calculation with contingency planning so that the set of "linear" plans represented by the conditional plan is robust accross all outcomes. Figure 5 illustrates the form of a conditional plan with the selected actions indicated.

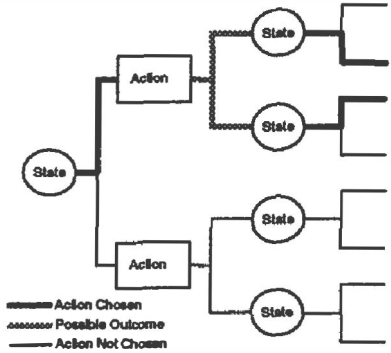

Figure 5. Conditional Plan

Of course, the exponential growth of the AND/OR tree with increasing depth makes an exhaustive search impractical in all but trivial domains. We will adopt the approach commonly used for minimax trees representing two-player, zero-sum, perfect information games [Nilsson, 1980]; specifically, we will employ a depth-limited search. By sacrificing exact computation of expected utility values and imposing a limited search horizon, the planning algorithm can be made reasonably efficient. Imposing a depth limit on the planning algorithm is very simple: we merely treat a state as terminal when it is at some fixed depth from the initial state in the AND/OR tree. The equations for expected utility can then be used to back expected utilities up to the initial state, as before. The depth limitation, used in this fashion, will have two impacts. First, the conditional plan will extend out only as far as the depth limit. Second, the expected utilities computed will be approximate, and therefore the actions chosen may not be optimal. The algorithm for computing conditional plans is formally described below.

```
procedure Expected_Utility_State(s,depth)
  if depth = depth_limit then
    return U_R(V(s))
  else
    actions := Generate_Actions(s)
    for each a in actions do
      EU_R(a|s) :=
              Expected_Utility_Action(a,s,depth)
      add to conditional plan the link from s to the
              a for which EU_R(a|s) is maximal
    return EU_R(a|s)
end

procedure Expected_Utility_Action(a,s,depth)
  sum := 0
  for each i do
    result_i(a,s) := s + Add_List(a,i) - Delete_List(a,i)
    sum := sum + P_a(result_i|s) *
            Expected_Utility_State(result_i(a,s), depth+1)
  return sum
end
```

### 4.1 Branch-and-Bound Planning Algorithm

Kumar and Kanal have presented a general formulation of branch-and-bound algorithms for AND/OR trees [Kumar, 1983]. In current work, we have shown that the planning algorithm presented above can be recast in such a branch-and-bound framework and its efficiency significantly improved by means of *pruning*. In particular, we can maintain a lower bound on the expected utility of an action (AND node), called *alpha*, defined as the highest current value of the state (OR node) ancestors of the action. We can then prune the action's children as soon as we know that its value will be less than or equal to *alpha*. The *alpha* bound is closely analogous to that used in the *alpha-beta* algorithm for minimax trees [Nilsson, 1980], which Kumar and Kanal have shown is also a special case of branch-and-bound. The branch-and-bound planning algorithm can be viewed as a form of *dominance-proving* planning, according to the terminology of [Wellman, 1990].

## 5 EXAMPLE DOMAIN: SLIPPERY BLOCKS WORLD

To test our formalism, we employed a simple domain for planning under uncertainty that is an extension of the blocks world. In the blocks world, blocks can be either on the table, on another block, or in the robot's hand. Every block is either clear or has a block on top of it. There is a robot arm that is capable of picking up any clear block and placing it either on the table or on another clear block. No more than one block may be placed directly on another block, but blocks may be stacked vertically to any height. There are four basic actions the robot arm can take, *pick-up*, *put-down*, *stack*, and *unstack*. Only one action may be performed at a time.

Our extended domain, the *slippery* blocks world, adds an element of uncertainty by making actions non-deterministic. In particular, each action that the robot arm performs has a chance of success, $P_a(result_{SC}|s)$, where $a$ is the type of action performed, and a probability of failure, $P_a(result_F|s)$. Failure leaves a state unchanged, i.e., $result_F(a,s) = s$. In our experiments, we chose for the sake of simplicity to make $P_a(result_{SC}|s)$ the same for all actions and states.

The value of a particular state is based on the arrangement of the blocks. Each block, $b$, has a worth $w(b)$ that is used to compute the value of the state, $s$. The value of the state is determined as follows.

$$v(s) = \sum_{b = \text{all blocks}} w(b)h(b)$$

where $h(b)$ is the height of the block above the table, e.g., $h(b) = 1$ for a block on the table, $h(b) = 2$ for a block on a block that is on the table, etc.

If the robot arm currently holds a block, it may *put-down* the block on the table or *stack* the block on any clear block. If the robot arm is empty it may *pick-up* any block from the table, or it may *unstack* any clear block. The decision operators may be described as shown in Figure 2, Figure 6, Figure 7 and Figure 8.

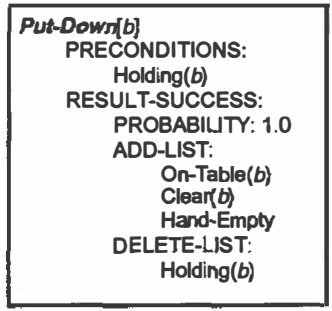

Figure 6. The Put-Down Operator

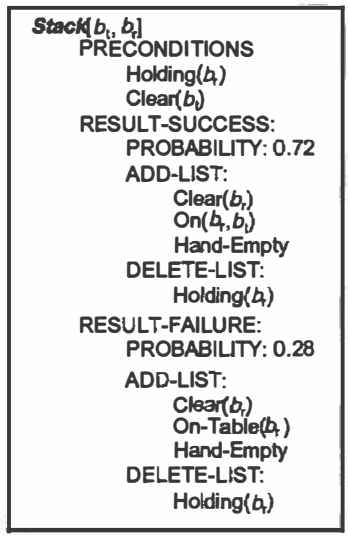

Figure 7. The Stack Operator

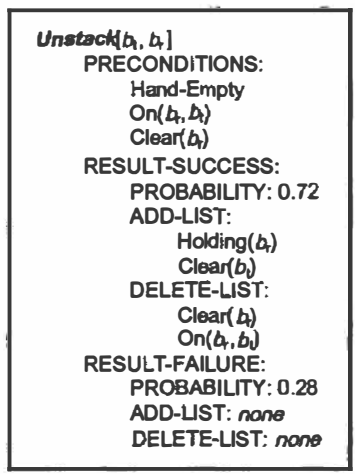

Figure 8. The Unstack Operator

### 5.1 Experimental Results

To obtain our experimental results, we employed Monte Carlo simulations of conditional plans generated using the algorithm of section 4. This involved traversing the conditional plan from the initial state, choosing the success or failure result of each action according to a fixed probability and then computing the value of the final state. We refer to the fixed probability used during simulation as the *execution* probability, as distinct from $P_a(result_{SC}|s)$. The planning algorithm was executed using a depth-limit of 6, $P_a(result_{SC}|s) = 0.72$, and an initial state as shown in Figure 9. The label of each block denote its worth. Conditional plans were created using robustness factors of 0.5 and 0.6.

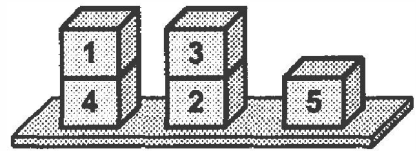

Figure 9. Initial State of Blocks in the Example

We first performed the Monte Carlo simulations 1000 times for execution probabilities of 0, 0.1, 0.2,... to 1.0. The results of these simulations are graphed in Figure 10.

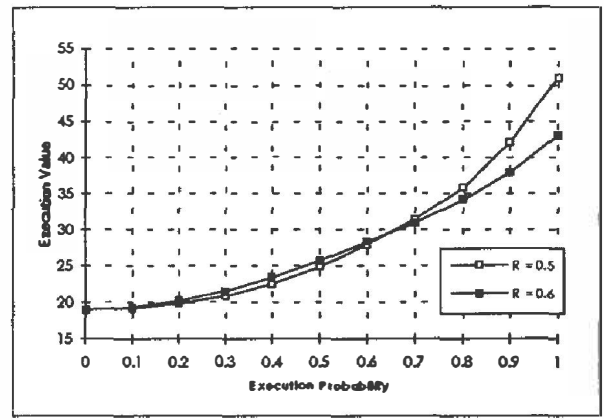

Figure 10. Graph of Execution Value vs. Execution Probability at Robustness Factors R = 0.5 and R = 0.6

We then ran 10000 Monte Carlo simulations of the two conditional plans, with the execution probability equal to $P_a(result_{SC}|s)$. We evaluated the final states of the simulations and produced a histogram of the results (see Figure 11). As shown in
Table 1 the plan created with a robustness of 0.6 had a lower mean value for the simulations, but also a lower standard deviation. Figure 12, created using the same data, shows the percentage of simulations which achieve at least a given value.

468  Pimentel and Brem

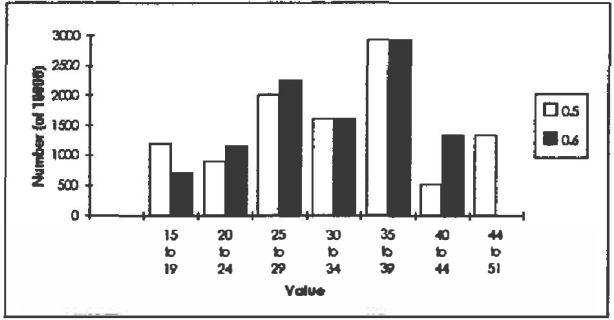

Figure 11. Histogram of Values for the Monte Carlo Simulations

Table 1. Mean and Standard Deviation for Robustness 0.5 and 0.6

|  | R = 0.5 | R = 0.6 |
|---|---|---|
| Mean | 32.06 | 31.50 |
| Standard Deviation | 9.67 | 6.82 |

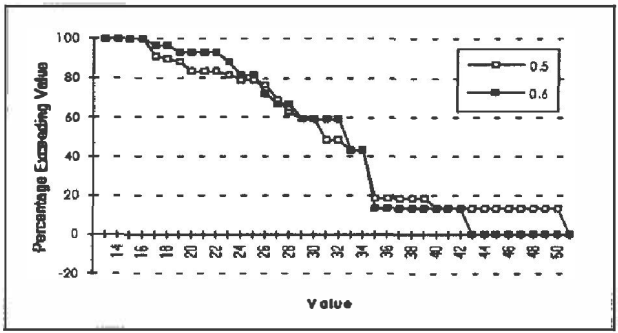

Figure 12. Percentage of Simulations Exceeding Given Value

Table 2 illustrates the difference between the two conditional plans by showing a single path through the plans in which all actions succeed.

Table 2. Success Path of Conditional Plans

|  | R = 0.5 | R = 0.6 |
|---|---|---|
| Action 1: | Pick-Up (1) | Pick-Up (5) |
| Action 2: | Stack (1, 3) | Stack (5, 1) |
| Action 3: | Pick-Up (5) | Pick-Up (3) |
| Action 4: | Stack (5, 1) | Stack (3, 5) |
| Action 5: | Pick-Up (4) | Pick-Up (2) |
| Action 6: | Stack (4, 5) | Stack (2, 3) |

Figure 13 shows the strategy employed by the plan with a robustness factor of 0.5. In Figure 13(a) the robot hand has placed block 1 on the 2-3 stack assuming that it will be able to put block 5 on block 1 next and block 4 on top. The major goal for the plan is to get blocks 5 and 4 as high as possible. This plan is a good strategy if and only if at least the first 4 actions shown on Table 2 are successfully accomplished. If only the first three actions are successful, then the system will get a score of 13. Figure 13(b) shows the best possible state that this conditional plan can generate. It has a score of 51, but all of the actions must be successfully executed to achieve this state.

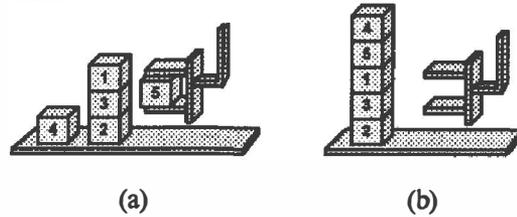

(a)            (b)

Figure 13. Two States in the Success Path of the Conditional Plan Generated using R = 0.5

Figure 14 shows the strategy employed by the plan with a robustness factor of 0.6. The planner, in this case, has a more pessimistic outlook. It immediately tries to move block 5 onto the 4-1 stack as shown in Figure 14(a). In this plan, block 5 will only be three blocks above the table, and therefore the best possible state that can be achieved is as shown in Figure 14(b). This state has a score of 43. The benefit of using this strategy is that it only requires two successful actions to move block 5 to its goal location. The worst possible score with block 5 in its goal location is as shown in Figure 14(c). This state has a score of 23.

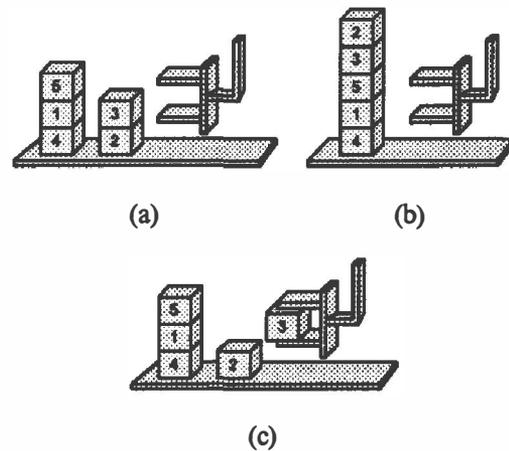

Figure 14. Three States in the Success Path of the Conditional Plan Generated using R = 0.6

These two examples show the effect of the robustness factor. The (R = 0.6) plan is less likely than the (R = 0.5) plan to result in a final state with a score less than 23. However, in contrast, the (R = 0.5) plan can achieve a higher optimal score than the (R = 0.6) plan. In fact, the expected value of the (R = 0.5) plan is higher than the (R = 0.6) plan. It would be appropriate to use the (R = 0.6) plan if the user wanted to "guarantee" that the final state will have a score greater than 20.

The robustness factor controls the amount of risk the planner is willing to take. For low robustness factors, the behavior of the planner is weighted more towards maximizing its expected value than ensuring at least a certain value will be achieved. In other words, the planner makes a trade-off between a higher expected value and a lower variance. A high robustness factor is important in domains where there is a significant likelihood of adverse occurrences: "It is better to pay for insurance so that, in the event of a misfortune, one will not be bankrupt."

## 6 RELATED WORK

The decision-theoretic principles employed in this planning are treated in detail by Keeney and Raiffa [Keeney, 1976]. The idea of applying decision theory to AI planning was first suggested by Feldman and Sproull in the late 1970's [Feldman, 1977]. However, the idea was not pursued by the AI planning community at that time. In the late 1980's, researchers began to examine ways of incorporating probabilistic information into planners [Kanazawa, 1989], and using decision theory to control search [Russell, 1989]. In a similar vein, Wellman and Doyle have advocated the explicit use of utility functions to guide the planning process [Wellman, 1991, 1992]. While building on these previous approaches, our work is the first to propose the use of a robustness factor to parameterize the utility function and to modulate the degree of risk-aversion sought by the planner.